\documentclass[12pt,a4paper,onecolumn]{IEEEtran}

\usepackage{algorithmic}
\usepackage{algorithm}
\usepackage[colorlinks=true,citecolor=magenta,urlcolor=blue]{hyperref}
\usepackage{cite}
\usepackage{graphicx,epstopdf}
\usepackage{rotating}
\usepackage{epsfig}
\usepackage[cmex10]{amsmath}
\usepackage{amssymb}
\usepackage{amsthm}
\usepackage{float}
\usepackage[tight,footnotesize]{subfigure}
\usepackage{array}
\usepackage{color}
\usepackage{multirow}
\ifCLASSINFOpdf
\else
\fi

\begin{document}

\title{Spatial Scaling of Satellite Soil Moisture using Temporal Correlations and Ensemble Learning} 
\author{Subit~Chakrabarti,~\IEEEmembership{Student~Member,~IEEE},~Jasmeet~Judge,~\IEEEmembership{Senior~Member,~IEEE},~Tara~Bongiovanni,~Anand~Rangarajan,~\IEEEmembership{Member,~IEEE},\\~Sanjay~Ranka,~\IEEEmembership{Fellow,~IEEE}.%
\thanks{A version of this manuscript was submitted to the IEEE\textsuperscript{\textregistered} Transactions on Geoscience and Remote Sensing.

This work was supported by the NASA-Terrestrial Hydrology Program (THP)-NNX13AD04G. The authors acknowledge computational resources and support provided by the University of Florida High-Performance Computing Center for all the model simulations conducted in this study.

S.~Chakrabarti, T.~Bongiovanni and J.~Judge are with the Center for Remote Sensing, 
Agricultural and Biological Engineering Department, 
Institute of Food and Agricultural Sciences, University of Florida, Gainesville, USA; A.~Rangarajan and S.~Ranka are with the Department of Computer \& Information Science \& Engineering, University of Florida, Gainesville.
E-mail: \href{mailto:subitc@ufl.edu}{subitc@ufl.edu}}}



\maketitle
\begin{abstract}
\boldmath
A novel algorithm is developed to downscale soil moisture (SM), obtained at satellite scales of 10-40 km by utilizing its temporal correlations to historical auxiliary data at finer scales. Including such correlations drastically reduces the size of the training set needed, accounts for time-lagged relationships, and enables downscaling even in the presence of short gaps in the auxiliary data. The algorithm is based upon bagged regression trees (BRT) and uses correlations between high-resolution remote sensing products and SM observations. The algorithm trains multiple regression trees and automatically chooses the trees that generate the best downscaled estimates. The algorithm was evaluated using a multi-scale synthetic dataset in north central Florida for two years, including two growing seasons of corn and one growing season of cotton per year. The time-averaged error across the region was found to be 0.01 $\mathrm{m}^3/\mathrm{m}^3$, with a standard deviation of 0.012 $\mathrm{m}^3/\mathrm{m}^3$ when 0.02\% of the data were used for training in addition to temporal correlations from the past seven days, and all available data from the past year. The maximum spatially averaged errors obtained using this algorithm in downscaled SM were 0.005 $\mathrm{m}^3/\mathrm{m}^3$, for pixels with cotton land-cover. When land surface temperature~(LST) on the day of downscaling was not included in the algorithm to simulate ``data gaps", the spatially averaged error increased minimally by 0.015 $\mathrm{m}^3/\mathrm{m}^3$ when LST is unavailable on the day of downscaling. The results indicate that the BRT-based algorithm provides high accuracy for downscaling SM using complex non-linear spatio-temporal correlations, under heterogeneous micro meteorological conditions.


\end{abstract}

\newpage

\section{Introduction}
%
%
%
%
Spatio-temporal distribution of soil moisture (SM) is highly variable and it significantly influences atmospheric and hydrological processes. Accurate SM information at spatial scales of $<$1~km is critical for applications such as agricultural drought monitoring, risk management, and productivity predictions, with major implications for food security and sustainability. Microwave observations such as those from the European Space Agency's Soil Moisture and Ocean Salinity (SMOS) mission and the NASA-Soil Moisture Active Passive (SMAP) mission, provides global observations of SM at spatial resolutions of 10-40 km every 2-3~days. These satellite-based SM obtained at coarse spatial resolutions need to be downscaled to about 1 km for applications in heterogeneous agricultural regions.

Most current downscaling algorithms are based upon linear unmixing algorithms, potentially leading to significant loss of structural information in the data~\cite{Chakrabarti2014}, especially under highly non-linear heterogeneous and dynamic conditions such as those in agricultural lands. Hierarchical models~\cite{Crow2002,Loew2008,Kim2002,Bindlish2002,Gebremichael2009,Manfreda2007} or empirical models based upon statistical~\cite{Piles2014,Ranney2015,Hsu2015,Ahmad2010,Liu2008,Chauhan2003,Merlin2013,Merlin2008,Merlin2013,Piles2011,Kim2012} relationships among SM and other remotely-sensed (RS) spatially cross-correlated data have been used for downscaling, assuming linear or quadratic relationships. The use of Bayesian descriptors of probability density functions (PDF) allows downscaling even in the presence of complex non-linear relationships between the coarse pixel and the multiple fine pixels. Recently, algorithms using higher order relationships have been developed to downscale SM using auxiliary high resolution RS products - for example, methods based upon Principle of Relevant Information (PRI)~\cite{Chakrabarti2014,Chakrabarti2013}, Support Vector Regression~\cite{Srivastava2013} and Copulas~\cite{verhoest2015}. Even though these methods result in significant improvement, extensive training is still required for low downscaling errors.


Current downscaling algorithms assume that all the augmented observations are available at the same time coarse SM is available, and cannot be applied in the presence of temporal data-gaps during cloud-cover, sensor recalibration exercises, solar flares, \textit{etc.} In addition, the relationship of some high resolution observations, such as precipitation, to soil moisture is discernible only over extended periods of time. Historical SM and its temporal correlations to other variables can be utilized to downscale SM in the presence of data gaps, as well as to minimize errors in downscaled SM due to time lagged relationships. These temporal correlations can also be leveraged for downscaling SM in regions where accurate characterization of the relationships between fine-resolution SM and auxiliary fine scale data inside a coarse pixel requires extensive training data of \textit{in-situ} SM. While implementation of downscaling algorithms using comprehensive training sets is realistic in data-rich agricultural regions in developed countries, the number of \textit{in-situ} stations sharply decreases in developing and underdeveloped countries. Augmenting the training set using historical observed data can increase the efficacy of downscaling algorithms. Significant gaps exist in utilizing historical RS and \textit{in-situ} data to implement downscaling algorithms involving implicit or explicit temporal prediction components.

Adding historical data to algorithms such as the PRI is impractical due to computational burden arising from non-stationarity, quick-switching physical regimes and fitting large number of hyper-parameters. Non-linear regression techniques can be used as they have higher-order prediction capabilities and can be parallelized for computational efficiency. However, these techniques tend to over-fit~\cite{Babyak2004} and need to be regularized to provide reliable results. Implicit regression methods are more suited to this problem as they are more efficient than explicit regression methods, such as PRI or non-linear regression, and they can be easily regularized. In this study, the most widely used implicit regression method, tree based regression~\cite{breiman1984}, is used to downscale soil moisture using historical information of correlated auxillary features. It is regularized using ensemble learning techniques~\cite{breiman1996}, that combine results of multiple models to reduce over-fitting.

The goal of this study is to understand the impact of incorporating historical information of SM in improving downscaled estimates in the presence of temporal data gaps or in data sparse regions. The primary objectives of this study are to 1)~develop an algorithm to downscale coarse scale SM using historical information of auxiliary variables and SM; 2)~implement this algorithm to downscale SM to 1~km using land surface temperature (LST), land cover (LC), leaf area index (LAI) and PPT; 3)~quantify the algorithm performance for application in data-sparse regions and in the presence of temporal data-gaps. In this study, a synthetic dataset~\cite{Nagarajan2012} will be used for the above objectives. 

Section~\ref{sec:Theory} describes the theoretical details of the downscaling framework. Section~\ref{Sec:Methodology} illustrates the steps for the implementation of the algorithm, presents the downscaling results for SM at 1~km and discusses the impact of both temporal training and temporal data gaps on the downscaled SM. Section~\ref{Sec:Conclusion} summarizes the results and concludes the paper.

\section{Theory}
\label{sec:Theory}

Downscaling, in general, is an ill-posed problem which results in a multiplicity of solutions after regression, requiring regularization. In this study, the downscaled SM is obtained using regression trees~(RT)~\cite{breiman1984}. To regularize the fine-scale SM estimates using RTs, a powerful non-linear and non-parametric ensemble learning technique called bootstrap aggregation (bagging)~\cite{breiman1996} is used. 

\subsection{Regression Trees}
\begin{figure}[t]
\centering\noindent
\centering\noindent\includegraphics[width=6 in]{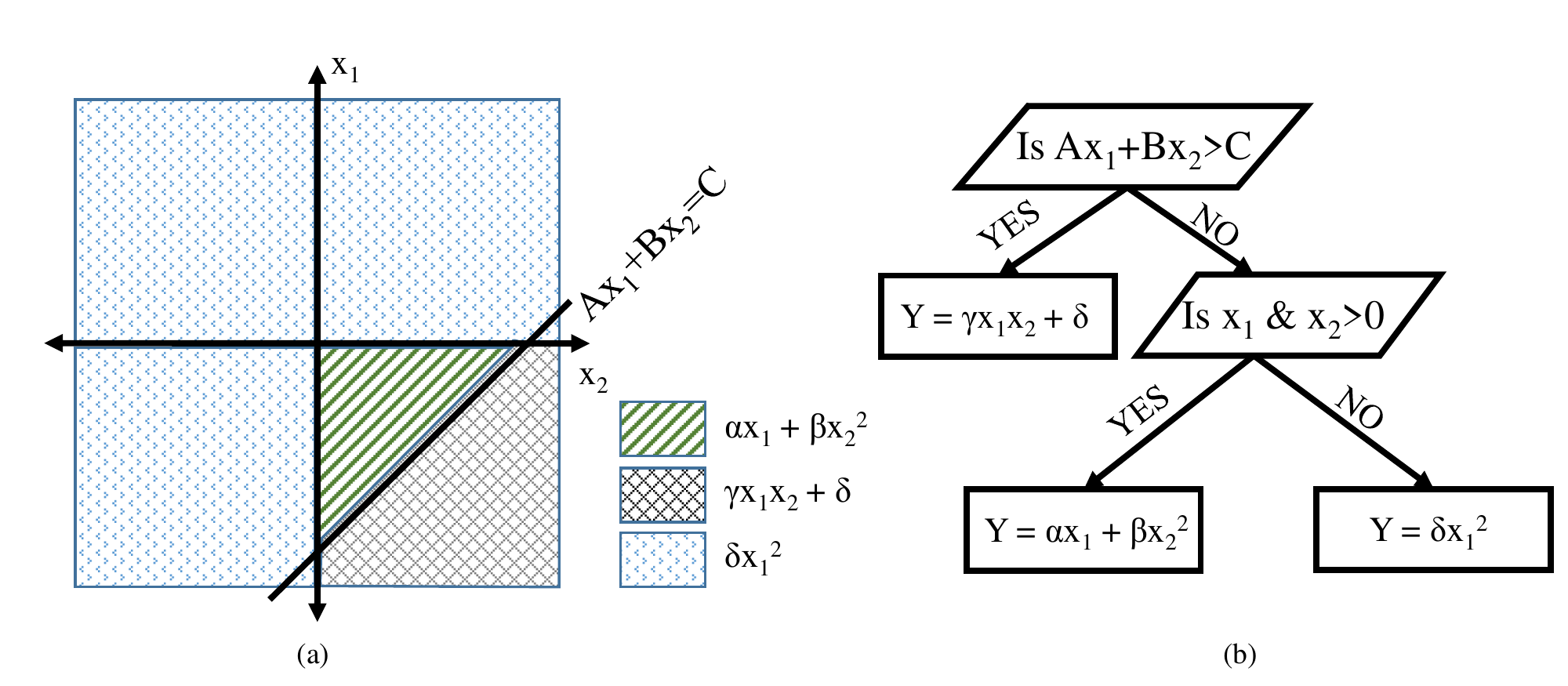}
\caption{(a) A two-dimensional vector space with 3 generative models, shown by shaded spaces, corresponding to 3 regression models and, (b) the associated regression tree which takes $x_1$ and $x_2$ as input and assigns the correct regression model corresponding to the vector space divisions. The parallelograms enclose partitioning rules and the rectangles enclose regression functions assigned to each partition.}
\label{fig:example}
\end{figure}

Linear or kernel regression are global methods, with a single predictive model used over the full range of the input data. However, heterogeneous and dynamic input variables, such has PPT and LC, interact with each other and exhibit highly non-linear correlations with SM making a single model sub-optimal. RTs~\cite{breiman1984} subdivide or partition the space of the inputs, such as auxiliary variables and coarse scale soil moisture, recursively and build simple regression models for each partition that maps to the required output, such as downscaled SM. Figure\footnote{All \hyperref[Figures]{Figures} and \hyperref[Tables]{Tables} are included at the end of the manuscript for clarity.}\addtocounter{footnote}{-1}\addtocounter{Hfootnote}{-1}~\ref{fig:example} shows an example of a single RT's ability to reconstruct a partitioned model space with simple non-linear models. The models themselves are shown for illustrative purposes. Because both the partitions within the input data space and the regression models can be non-linear, the RT method can approximately learn any mapping function between the input and output data-sets. If the number of partitions is too few, the regression models would be complex and fail to capture the structure of the data, while if the number of partitions is too many, the models would over-fit and perform poorly in the presence of noise in the input data. 

An algorithm to construct an RT consists of a heuristic, used to recursively select which input variable to use to partition the input space for each level of the tree, henceforth called the partitioned variable, and the stopping criterion, that specifies when each partition reaches the optimal size. In this study, the partitioned variable for each level of the tree is decided based upon the  sum of squared errors on the validation set after regression. For each tree, $t$, the prediction error, $E(t)$, is defined as,

\begin{align}
E(t) &= \sum_{\forall l \in \mathrm{Leaf}\{t\}}  \sum_{\forall i \in l}\left( \frac{1}{n_L} \sum_{\forall i \in l} y_i - y_i \right) \nonumber \\
& = \sum_{\forall l \in \mathrm{Leaf}\{T\}}n_l \left( \sum_{\forall i \in l}\left( \frac{1}{n_l^2} \sum_{\forall i \in l} y_i - y_i \right)\right) \nonumber \\
& = \sum_{\forall l \in \mathrm{Leaf}\{t\}}n_L \cdot \mathrm{Var}(l) 
\label{eq:Error}
\end{align}
where $y = \hat{f}(x)$ is the predicted variable, $\mathrm{Var}(l)$ is the variance in the set of possible predictions for leaf, $l$. The optimal partition results in minimization of the prediction error, E, on the validation set. In this study, the stopping criterion is achieved when additional splits decrease $E(t)$ by $<$0.01 $\mathrm{m}^3/\mathrm{m}^3$ SM.

\subsection{Bootstrap Aggregation and Ensemble Pruning}

A single regression tree, as described above, is prone to over-fitting the data. Bootstrap Aggregation or Bagging~\cite{breiman1996} is an ensemble learning technique that re-samples the original training set uniformly to create multiple new data-sets of the same size as the original data-set. Each bootstrapped sample is fitted using as single RT, resulting in the same number of RTs as the number of datasets, and the mean of the outputs as the final output. This procedure has been shown to improve the stability of individual trees which might over-fit the data due to their non-parametric nature~\cite{breiman1996}.

For computational efficiency, ensemble pruning is employed. This technique uses the ensemble of models generated by bagging and reduces the number of models by weeding models that perform poorly on a validation set, while maintaining the diversity of models. The $L^1$-constrained fitting for statistics and data mining (LASSO)~\cite{tibshirani1996} algorithm has been widely used for this purpose. For $K$ RTs, the mean square loss~\cite{Narsky2013}, $E_{overall}$, for a validation set $\{x_{1}\ldots x_{N}\}$ is,

\begin{align}
E_{overall} = \sum_{n=1}^N w_n \left( f(x_n) - \sum_{k=1}^K \alpha_k\hat{f}_k(x_n)\right)^2 + \beta \sum_{k=1}^K \alpha_k
\label{eq:true}
\end{align}
where $f(x_n)$ is the true value of the function, $\hat{f}_k(x_n)$ is the estimated value of the function by the $k^{th}$ RT, $\beta$ is the regularization parameter, $w_n$ are the weights to be learnt and $\alpha_k$ is the learner coefficients. Learning the weights then is equivalent to solving the following:
\begin{align}
\mathrm{min}_{\alpha} \sum_{n=1}^N w_n \left( f(x_n) - \sum_{k=1}^K \alpha_k\hat{f}_k(x_n)\right)^2 \textrm{such that} \sum_{k=1}^K \alpha_k \leq \frac{1}{\lambda}
\label{eq:cost}
\end{align}
where $\lambda$ is a constant that regularizes the weights, $\alpha_k$. The higher $\lambda$ is, the more trees are selected after ensemble pruning. The solution provides the optimum weights for each weak learner. The $l$ learners with the highest weights are then selected. This technique naturally avoids over-fitting and allows for extremely powerful generalization.

\subsection{Algorithm Summary}

\begin{figure}[t]
\centering\noindent
\centering\noindent\includegraphics[width=6 in]{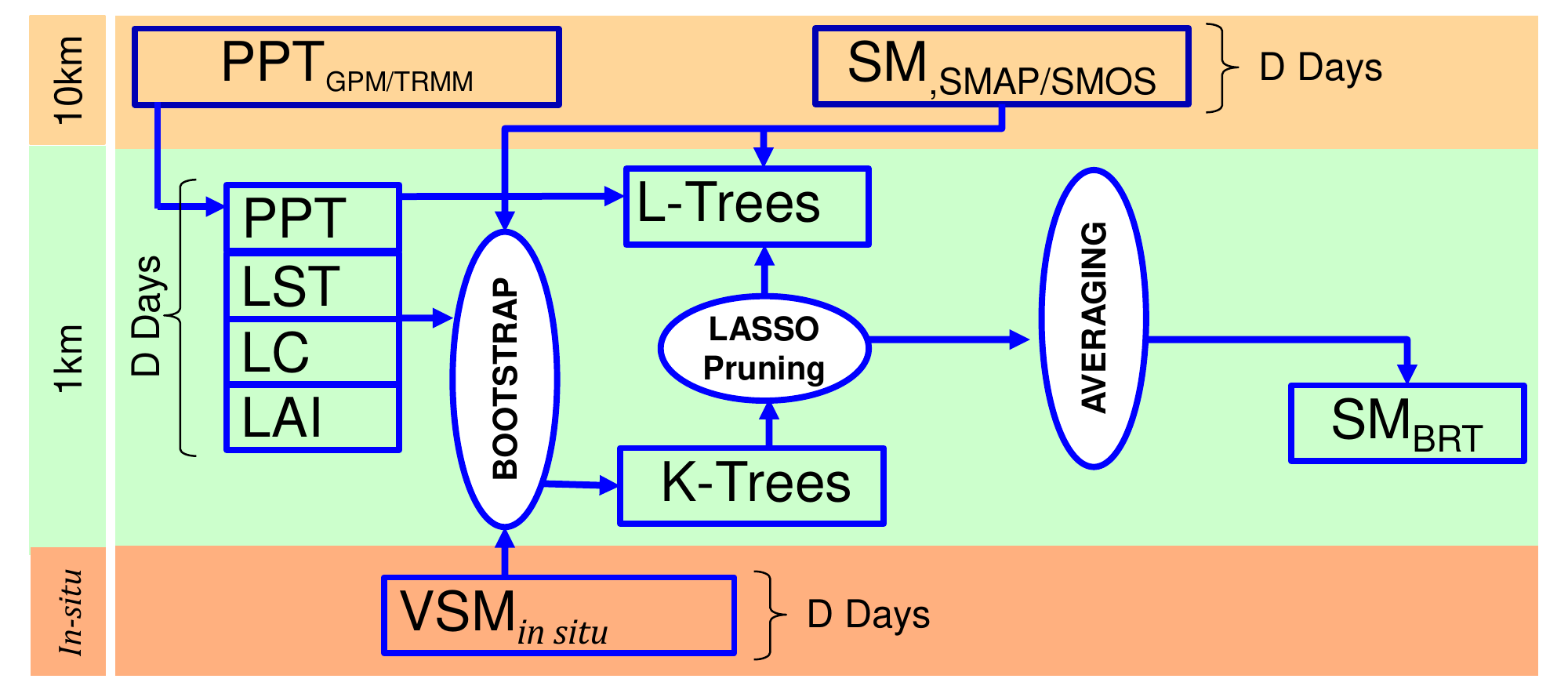}
\caption{Flowchart showing the boosted regression trees (BRT) downscaling method.}
\label{fig:flow}
\end{figure}

The overall method of bagged regression trees~(BRT) is shown in Figure~\ref{fig:flow}. For incorporating historical information into the algorithm a spatio-temporal training data-set is created by augmenting current \textit{in-situ} SM and axillary data with data from the prior days. For each LC, separate replicates of the spatio-temporal dataset is created using the bootstrapping method. An equal number of regression trees as the replicates are grown for each bootstrapped~\cite{breiman1996} version of the data, and LASSO-pruning is used to separate out trees that best describe the data, as described in Section II. Finally, the coarse SM is bootstrapped and regressed using the selected trees and the averaged output of these is the downscaled SM. 

\section{Experimental Dataset}

\begin{figure}[t]
\centering\noindent
\includegraphics[width=3.5 in ,keepaspectratio=true]{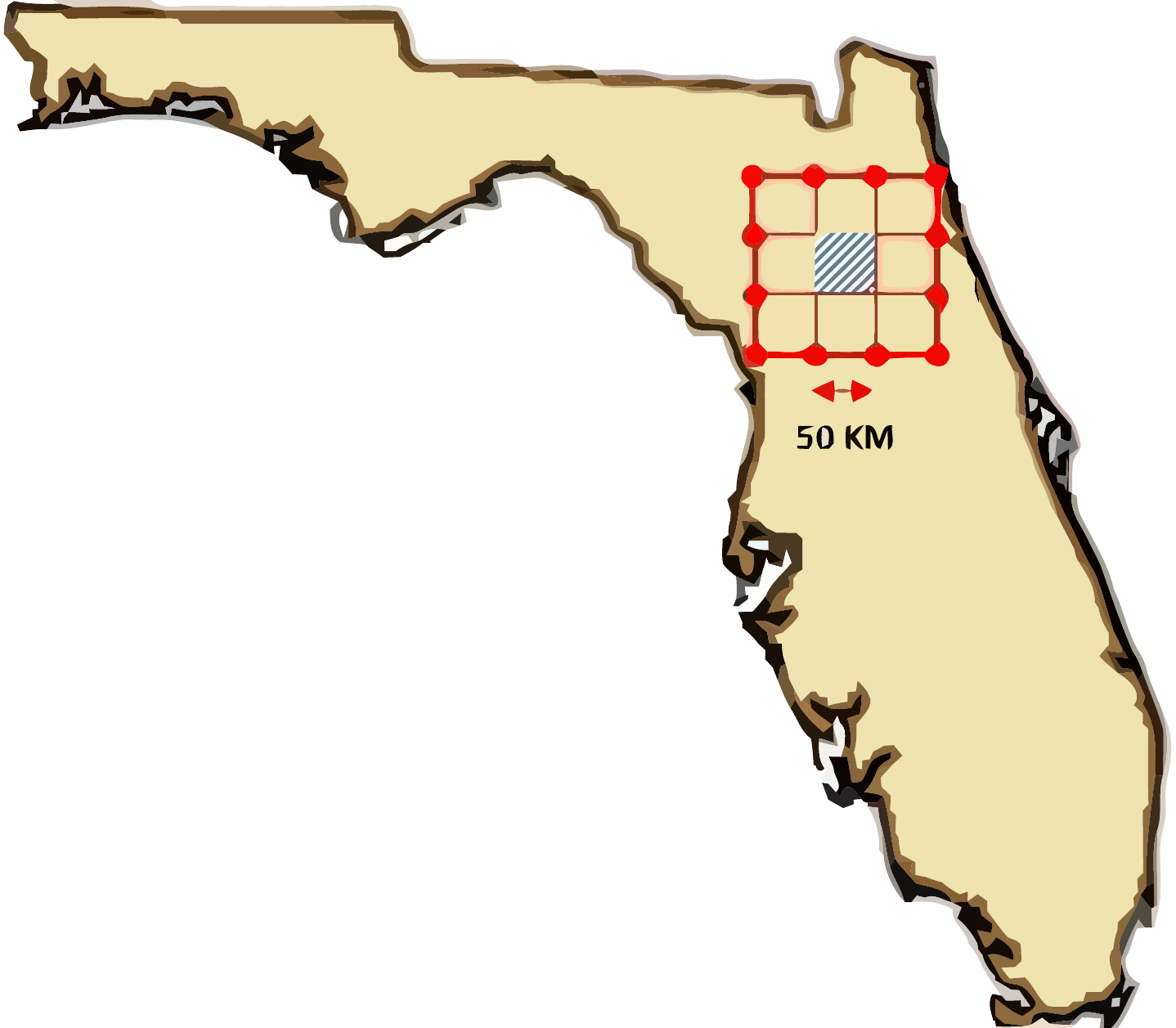}
\caption{Study region in North Central Florida. LSP-DSSAT-MB simulations were performed over the shaded $50\times50$ km$^2$ region.}
\label{fig:studysite}
\end{figure}

The downscaling  algorithm in the study used a dataset from the simulation framework consisting of a soil-vegetation-atmosphere transfer model, the Land Surface Process (LSP) model, coupled with a crop growth model, the Decision Support System for Agrotechnology Transfer (DSSAT) model, described in \cite{Nagarajan2012}. A $50\times50$ km$^2$ region, equivalent to approximately 25 SMAP pixels with a spatial resolution 9~km/pixel was chosen in North Central Florida (see Figure~\ref{fig:studysite}) for the simulations. The region encompassed the UF/IFAS Plant Science Research and Education Unit, Citra, FL, where a series of season-long field experiments, called the Microwave, Water and Energy Balance Experiments (MicroWEXs), have been conducted for various agricultural land covers over the last decade~\cite{Bongiovanni2009,Casanova2006,Lin2004}.  Simulated observations of SM, LST $\&$ LAI were generated at 200 m for two years, from January 1, 2007 through December 31, 2008. Topographic features, such as slope, constant as the region is typically characterized by flat and smooth terrains with no run-off due to soils with high sand content. The soil properties were assumed constant over the study region. 

\begin{figure}
\centering\noindent\includegraphics[width=7 in ,keepaspectratio=true,clip]{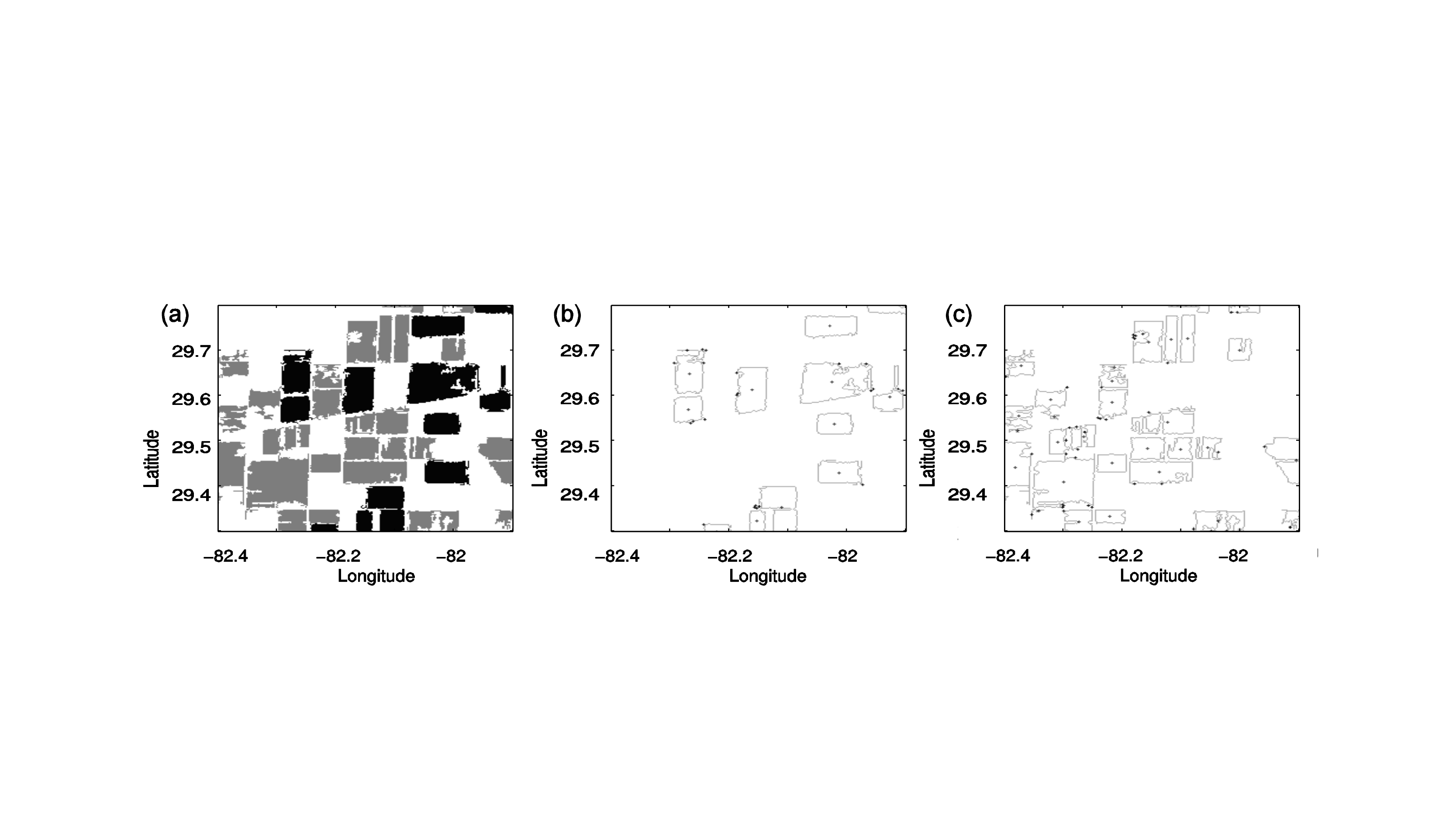}
\caption{(a) Land cover at 200m during cotton and corn seasons. White, gray, and black shades represent 
baresoil, cotton, and sweet-corn regions, respectively. Homogeneous crop fields along with centers for (b) sweet-corn and (c) cotton.}
\label{fig:landcover}
\end{figure}

For the LSP-DSSAT model simulations, fifteen-minute observations of PPT, relative humidity, air temperature, downwelling solar radiation, and wind speed were obtained from eight Florida Automated Weather Network (FAWN) stations located within the study region. The observations were spatially interpolated using splines to generate the meteorological forcings at 200 m. Long-wave radiation was estimated following~\cite{Brutsaert1975}. The model simulations were performed over each contiguous homogeneous region of sweet-corn, bare soil, and cotton, as shown in Figure~\ref{fig:landcover}, rather than all the pixels, to reduce computation time. A realization of the LSP-DSSAT model was used to simulate SM, LST and LAI at the centroid of each homogeneous region, using the corresponding crop module within DSSAT. The model simulations were performed using the 200 m forcings at the centroid, as shown in Figure~\ref{fig:landcover}. 

The model simulations at 200 m were spatially averaged to obtain PPT, LST, LAI, and SM at 1 and 10km. Linear averaging is typically sufficient to illustrate the effects of resolution degradation~\cite{Crow2003}. To simulate rain-fed systems, all the water input from both precipitation and irrigation were combined together, and the ``PPT" in this study represents these combined values. 

The simulation period in both the years consisted of two growing seasons of sweet corn and one season of cotton, as shown in Table\footnote{All \hyperref[Figures]{Figures} and \hyperref[Tables]{Tables} are included at the end of the manuscript for clarity.}\addtocounter{footnote}{-1}\addtocounter{Hfootnote}{-1}~\ref{tab:crops}. The LST, PPT, and LAI observations at 1~km, and the SM observations at 10~km were obtained by adding white Gaussian noise to the model simulations account for satellite observation errors, instrument measurement errors, and micro-meteorological variability, following \cite{Huang2008b,Privette2002,Crow2002}. The errors added had zero mean and standard deviations of 5K, 1 mm/hour, 0.1 and 0.02 $\mathrm{m}^3/\mathrm{m}^3$ for LST, PPT, LAI at 1km and SM at 10km, respectively, similar to~\cite{Chakrabarti2015} and~\cite{Chakrabarti2015b} algorithms.

\section{Methodology and Results}
\label{Sec:Methodology}
The inputs to the BRT method are LST, PPT, LAI, and LC at 1~km. The first part of the study, described in Section~\ref{Sec:BRTSPACE}, used only spatial correlations to downscale SM, similar to the current downscaling studies~\cite{Kim2012,Chakrabarti2015}. Two scenarios are implemented and compared: The first in which randomly selected 33\% of the dataset, or 750 out of 2500 pixels, are used for training (BRT$_\mathrm{750}$), similar to other training-set based downscaling algorithms~\cite{Ghosh2007,Chakrabarti2015,Chakrabarti2015b}. And second, in which randomly selected $0.02\%$ of the dataset, or 30 pixels, are used for training (BRT$_\mathrm{30}$). The second scenario represents realistic situations for data poor regions in developing countries. Both the BRT$_\mathrm{750}$ and BRT$_\mathrm{30}$ algorithms are implemented using the SM and auxiliary variables from Jan 1 through Dec 31, 2008. 

In the second component of the study, described in Section~\ref{Sec:BRTTEMP}, the BRT-based downscaling is modified and extended to utilize temporal correlations, in-addition to the spatial correlations (BRT$_\mathrm{st}$). This method is non-parametric and thus, can be easily extended to include temporal training The BRT$_\mathrm{st}$ algorithm is trained using $0.02\%$ of the dataset in space, and all the prior data available for those pixels, to provide temporal correlations. In this study, the prior data consists of all the data available from a year prior to the day of downscaling in 2008. 

In addition to handling spatial data sparsity, the BRT$_\mathrm{st}$ algorithm can be used to downscale SM in the presence of data-gaps. In the third component of the study, the effects of data-gaps in LST are investigated. Remotely-sensed LST, which is usually retrieved from the MODIS instrument on-board NASA Aqua and Terra satellites, is the most likely to be unavailable due to cloud gaps~\cite{Frey2013}. The near real-time LST products, which we envision will be used for this downscaling algorithm, are additionally affected by gaps caused due to solar gaps and imprefect satellite transmissions. In this study, the effect of data-gaps in LST is simulated by withholding LST data for the entire region on the day of downscaling, and on prior days, from the input data-set of the BRT$_\mathrm{st}$ algorithm. 


\subsection{BRT-based Downscaling with Spatial Correlations}
\label{Sec:BRTSPACE}
For downscaling based on only spatial correlations, the regression trees are trained using the auxiliary data-set on each day that the coarse SM is available. In Equation~\ref{eq:Error},~\ref{eq:true}, and \ref{eq:cost} from Section~\ref{sec:Theory}, 
\begin{align}
f^t(x^t_n)&=\mathrm{SM^t_{n,insitu}} \nonumber \\ x^t_n &= (\mathrm{LST}_n^\mathrm{1 km}, \mathrm{PPT}_n^\mathrm{1 km}, \mathrm{LAI}_n^\mathrm{1 km}, \mathrm{LC}_n^\mathrm{1 km}, \mathrm{SM}_n^\mathrm{10 km}, \mathrm{X}_n, \mathrm{Y}_n)^t \nonumber \\ \hat{f}^t(x^t_n) &= \mathrm{SM_{n,downscaled}^t}
\end{align}
where $X_n$ and $Y_n$ are the x and y coordinates of the $n^{th}$ pixel, superscript $t$ denotes the time index (in DoY) and $n \in[1,2500]$.  The average root mean square errors~(RMSE) for the BRT$_\mathrm{750}$ method for three land-covers are shown in Table~\ref{tab:pritable}. The errors for both vegetated and bare-soil pixels is very low, with bare-soil pixel having a higher mean error due to effects of fractional land-cover in the boundary pixels and crop remnants after harvest. The PRI~\cite{Chakrabarti2015} method also results in errors close to 0 under this conditions, with a similar training set size. The BRT$_\mathrm{30}$ algorithm produces errors with a mean of 0.067 $m^3/m^3$ and a standard deviation of 0.051 $m^3/m^3$, similar to the other algorithms that use only spatial correlations. This is expected since the spatial correlations in such low volume training sets aren't strong enough to yield definitive relationships that can be utilized to downscale SM. 
 
\subsection{BRT-based Downscaling with Spatio-Temporal Training}
\label{Sec:BRTTEMP}

\begin{figure}[t]
\centering\noindent
\includegraphics[width=7 in ,keepaspectratio=true]{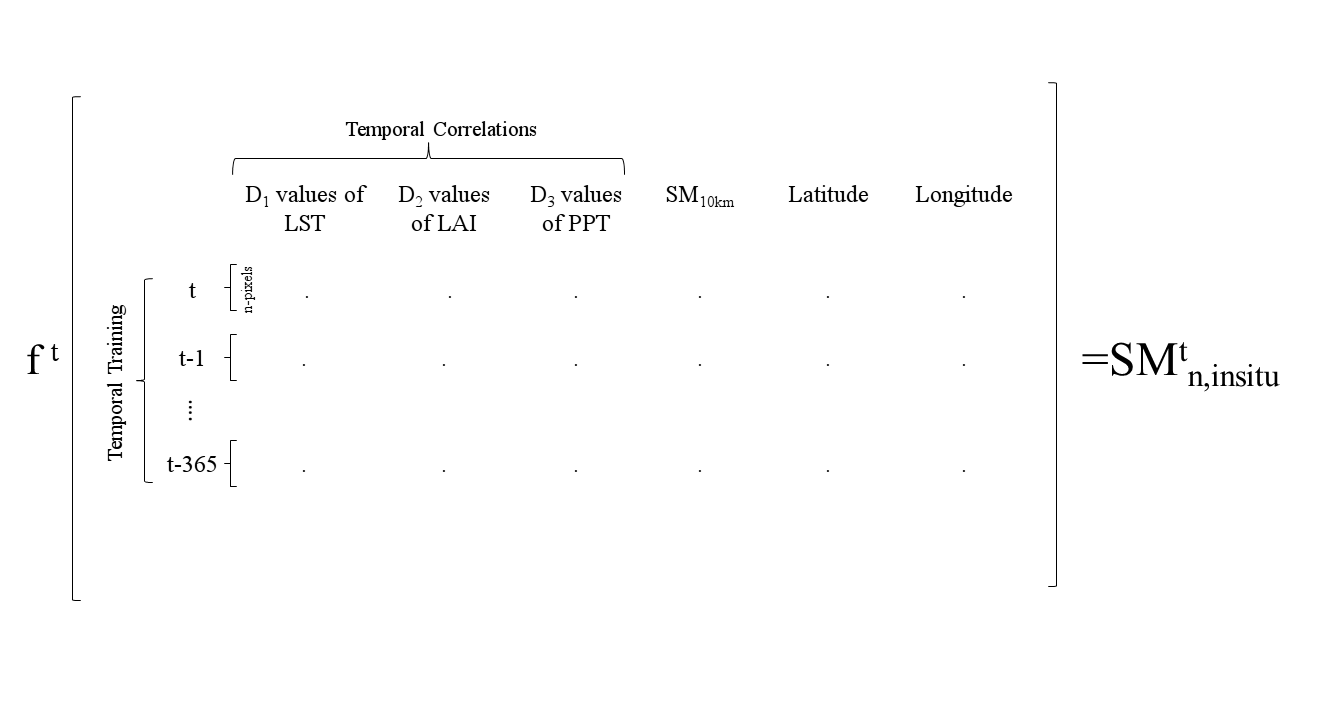}
\caption{Complete feature vector for the bagged regression trees algorithm with spatio-temporal training.}
\label{fig:feature}
\end{figure}

\begin{figure}
\centering\noindent\includegraphics[width=3.5 in ,keepaspectratio=true,clip]{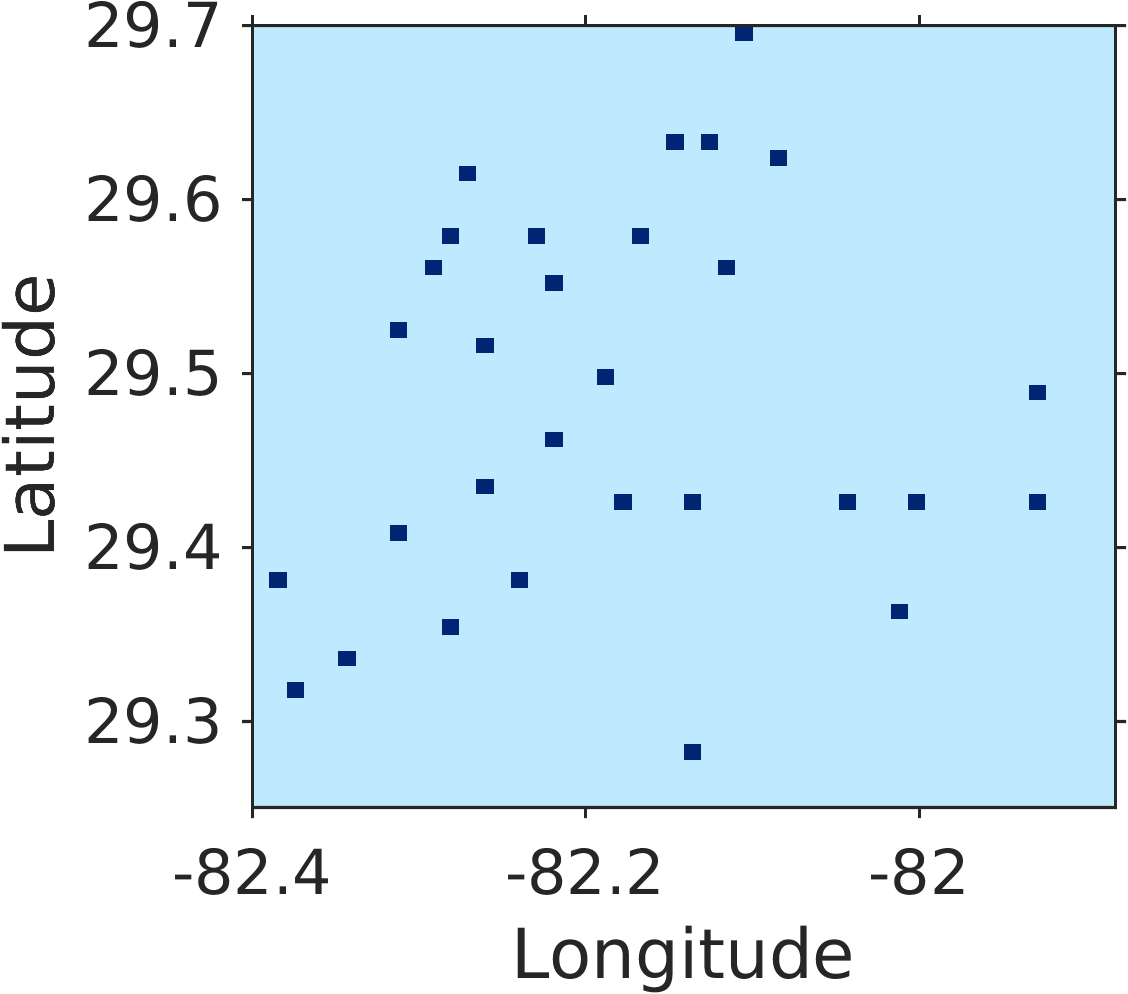}
\caption{30 Training locations within the study area for the bagged regression trees algorithm with spatio-temporal training.}
\label{fig:training_loc}
\end{figure}

For the BRT$_\mathrm{st}$ algorithm, a regression tree is trained when coarse-scale SM is available, using historical information of LST, LAI and PPT from the prior 365 days for 0.02\% of the pixels. Thus, 
\begin{align}
f^{t}(x_{nt})=&\mathrm{SM_{nt,insitu}} \nonumber \\ x_{nt} = &([\mathrm{LST}_{(t-D_1)n}^\mathrm{1 km}, \mathrm{LST}_{(t-D_1+1)n}^\mathrm{1 km}, \ldots , \mathrm{LST1}_{tn}^\mathrm{1 km}],[\mathrm{LAI}_{(t-D_2)n}^\mathrm{1 km}, \mathrm{LAI}_{(t-D_2+1)n}^\mathrm{1 km}, \ldots , \mathrm{LAI}_{tn}^\mathrm{1 km}], \nonumber \\ &[\mathrm{PPT}_{(t-D_3)n}^\mathrm{10 km}, \mathrm{PPT}_{(t-D_3+1)n}^\mathrm{10 km}, \ldots , \mathrm{PPT}_{tn}^\mathrm{1 km}], \mathrm{SM}_{tn}^\mathrm{10 km}, \mathrm{X}_{tn}^\mathrm{10 km}, \mathrm{Y}_{tn}^\mathrm{10 km}]) \nonumber \\ \hat{f}^{t}(x_{nt}) = &\mathrm{SM_{nt,downscaled}}
\label{eq:temporal}
\end{align}
where $n\in [1,2500]$ is the space-index, $t\in [1,365]$ is the time index and $D_1$, $D_2$, and $D_3$ are the number of values of LST, LAI and PPT to include in the feature vector from previous days, respectively. For training purposes, LAI, available every 7 days, is assumed to be constant between consecutive observations, while LC, PPT and LST are available every day. The feature vector for downscaling SM is described in more detail in Figure~\ref{fig:feature}. The 30 pixels, $n$ in Figure~\ref{fig:feature}, are randomly selected to serve as the training set. For each time-index $t$, the $D_1$, $D_2$, and $D_3$ observations of LST, LAI and PPT, respectively, are included as the columns of the feature matrix that enable the algorithm to use temporal correlations and capture spatial soil-moisture variability. The range of the time index, $t$, in the rows of the feature matrix, ranging from $t-1$ to $t-365$ indicates that a moving window of historical auxiliary and \textit{in-situ} data from the past one year is used to augment the training data-set. For example, to downscale on DOY 47 in 2008, all the datasets from DOY 47, 2007 through DOY 47, 2008 for the 30 pixels selected randomly prior to the algorithm implementation were used for training, with a moving window of  $D_1$, $D_2$, and $D_3$ values of LST, LAI, and PPT, respectively, included for each day of training. The 30 locations in the study area used for training spatio-temporally are shown in Figure~\ref{fig:training_loc}.



\begin{figure}
\centering\noindent\includegraphics[width=7 in ,keepaspectratio=true,clip]{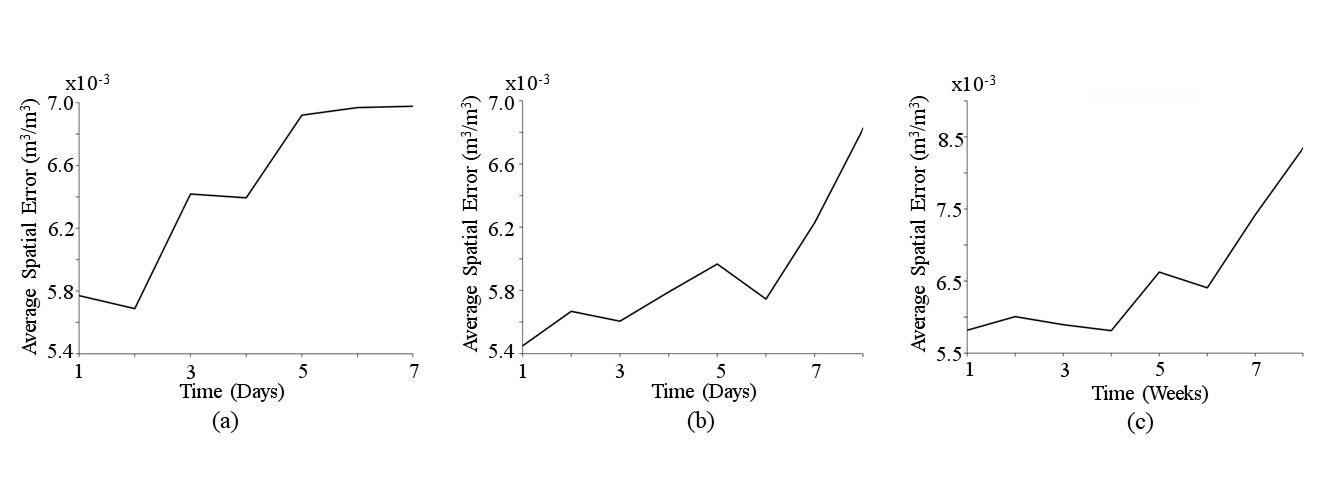}
\caption{Spatially averaged Error in downscaled soil moisture vs. (a) time-lag of observation for land surface temperature, (b) time-lag of observation for precipitation, and (c) time-Lag of observation for leaf area index.}
\label{fig:LST}
\end{figure}

The spatially-averaged errors in downscaling were compared for different values of $D_1$, $D_2$, and $D_3$, to understand the impact of utilizing temporal correlations. These errors were also averaged for 122 days in the year 2008. These spatio-temporally averaged errors, while underestimating the possible errors due to downscaling in areas with highly varying land-cover and/or meteorological conditions, provides a reasonable estimate of the upper bound of how much historical data to include, beyond which the variables become almost uncorrelated to the current values of SM. Figures 5(a)-(c) show these spatio-temporal errors for different values of $D_1$, $D_2$, and $D_3$. The errors when previous values of LST, PPT, and LAI are used are shown in Figures~\ref{fig:LST}. For LST and LAI, the spatially averaged error, increases significantly after 7 days and 7 weeks, respectively. This indicates that including additional LST information from earlier than 7 days or including additional LAI information from earlier than 6 weeks does not provide any added value to the downscaled estimates. The spatially averaged errors, for PPT, stabilizes after 5 days. However, the effect of PPT on SM is highly non-linear and is a function of the magnitude of and time-delay from the last rainfall event. In general, if we do not consider spurious correlations, such as those on the third day in Figure 5(b), PPT does not have a noticeable effect on SM beyond a week's lag, even for the rain-fall events, as high as 3.6 mm/hr. For the ease of implementation, seven prior values of PPT, LST and LAI data, an overestimate for $D_1$, $D_2$, and $D_3$ equal to 7, were added to the feature vector  ($x_{nt}$ in Equation 5). 

\begin{figure}
\centering\noindent\includegraphics[width=\textwidth]{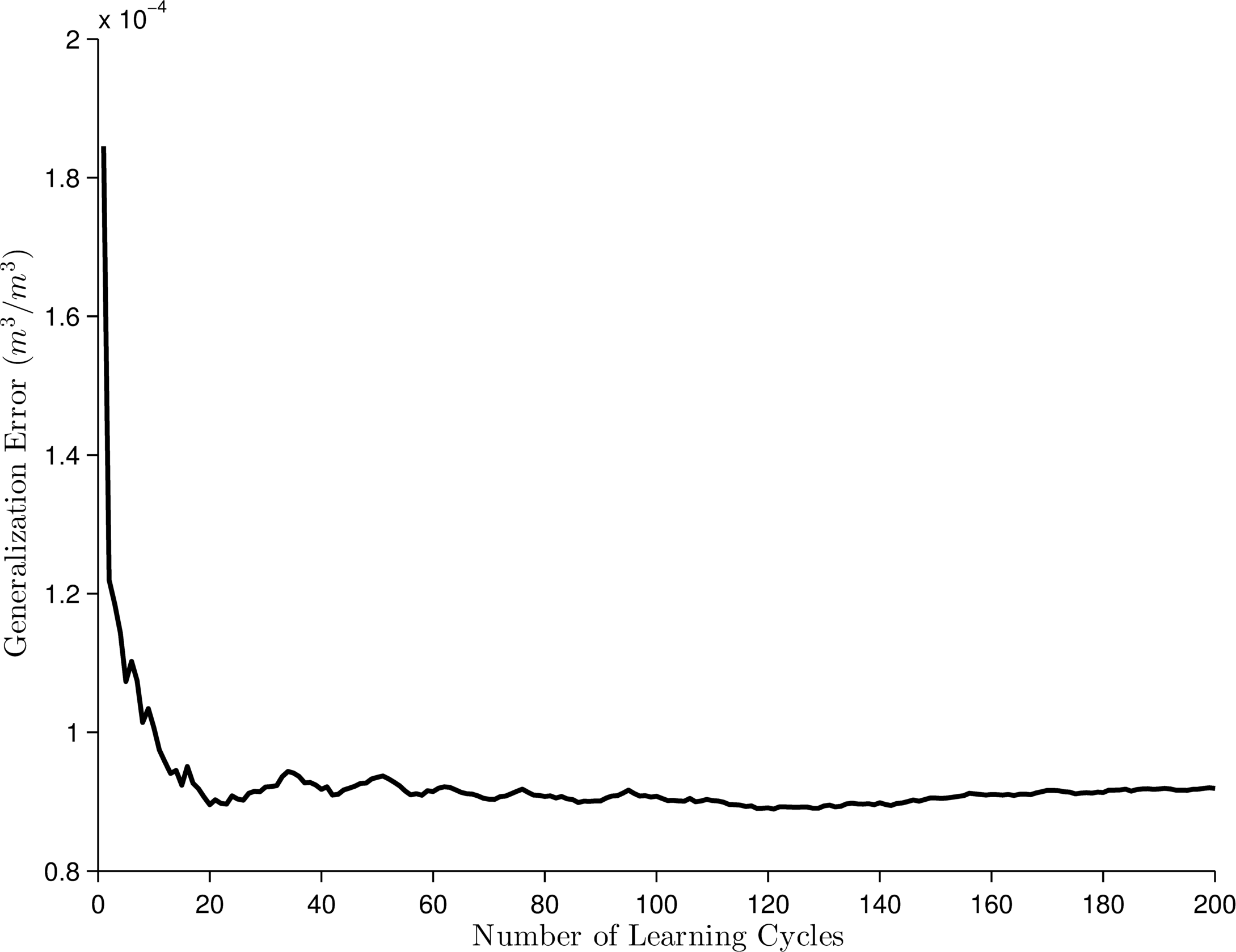}
\caption{Generalization error vs. no. of learning cycles for the bagged regression trees algorithm with spatio-temporal training.}
\label{fig:tree_error}
\end{figure}

\begin{figure}
\centering\noindent\includegraphics[width=0.8\textwidth]{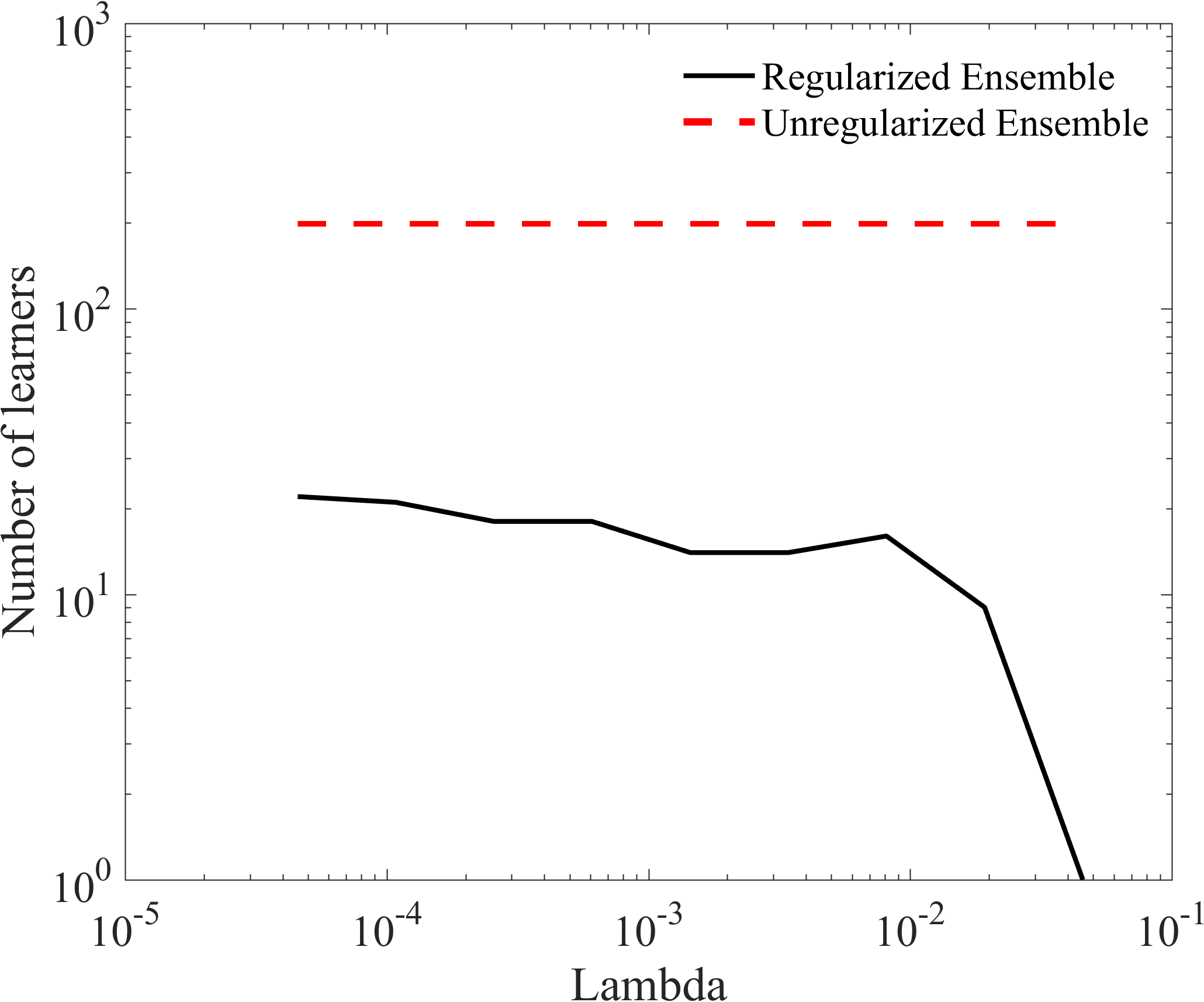}
\caption{No. of learners vs. lambda for the bagged regression trees algorithm with spatio-temporal training.}
\label{fig:model_error_2}
\end{figure}

\begin{figure}
\centering\noindent\includegraphics[width=0.8\textwidth]{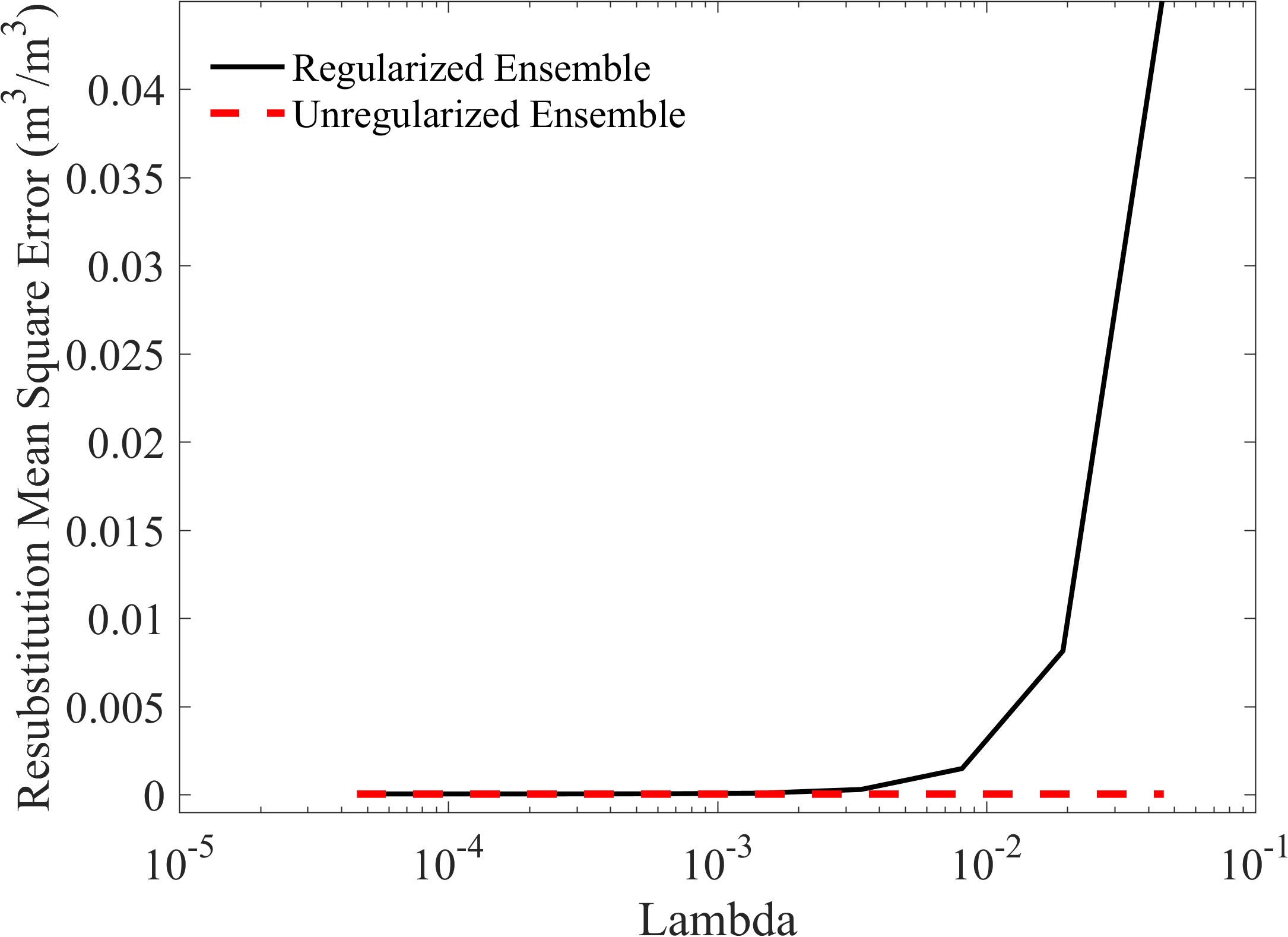}
\caption{Re-substitution Error vs. Lambda for the bagged regression trees algorithm with spatio-temporal training.}
\label{fig:model_error}
\end{figure}

The sensitivity of the BRT$_\mathrm{st}$ algorithm to the number of trees ($K$ in Equation 2) is investigated by dividing the training set into 10 equal parts randomly, and using 9 parts for training and one part for evaluating the algorithm. This methodology, known as 10-fold cross-validation, is repeated with different randomly selected partitions to approximate the errors that the BRT$_\mathrm{st}$ algorithm would incur on an average. Figure~\ref{fig:tree_error} shows the 10-fold cross-validation error as a function of the number of regression trees trained. Beyond 50 trees, the estimated error in downscaling decreases by only 0.05 $\mathrm{m}^3/\mathrm{m}^3$, which is an inconsequential amount compared to the extra computational time, of about 10 minutes for each day, incurred for each additional tree. Thus, in this study, 50 trees were used to estimate the downscaling function. To decrease over-fitting, after  the number of trees in the model is decreased even further after training through regularization, as shown in Figure~\ref{fig:model_error_2}. The lambda value of $2\times 10^{-2}$ is used which reduces the number of active trees needed to 20 for an average increase in error of 0.003 $\mathrm{m}^3/\mathrm{m}^3$ as shown in Figure~\ref{fig:model_error}. While applying the trained regression tree ensemble for downscaling, decreasing the number of trees from 50 to 20 reduces the computational time marginally by about 5 minutes per day. This reduction is expected to be much more significant when downscaling for multiple seasons for higher number of pixels.

The spatially averaged errors in SM during the year 2008 are $\leq$ 0.01 $\mathrm{m}^3/\mathrm{m}^3$, with the highest errors, of about 0.023 $\mathrm{m}^3/\mathrm{m}^3$ during simultaneous corn and cotton land-covers as well as late season bare soil due to remnant crops. Table~\ref{tab:pritable} shows the RMSEs for the BRT$_\mathrm{st}$ algorithm, for various land covers, as shown in Figure 13. The RMSE for the BRT$_\mathrm{st}$ method is higher than for the BRT$_\mathrm{750}$ algorithm by about 0.018 $\mathrm{m}^3/\mathrm{m}^3$. This is expected as the training set used by the BRT$_\mathrm{750}$ algorithm is 24 times larger in volume. The marginal RMSE increase in the SM downscaled using the BRT$_\mathrm{st}$ algorithm is justified considering the significant advantages from a smaller training data-set. 

Five days were selected from the season to understand the effect of the heterogeneity in inputs on the error in disaggregated SM.  Variabilities in precipitation, ranging from uniformly wet to uniformly dry, and in land cover, ranging from bare soil to vegetated with both cotton and sweetcorn, were used as criteria for selecting the days, as shown in Table~\ref{tab:conditions} and Figure~\ref{fig:daysinputs}. For the five selected days as shown in Table~\ref{tab:conditions}, the true SM and the downscaled SM are shown in Figures~\ref{fig:39}-\ref{fig:222}. Figures~\ref{fig:39} and~\ref{fig:354} shown the downscaled SM for early and late-season bare soil conditions. For the bare soil pixels the maximum error is 0.03 $\mathrm{m}^3/\mathrm{m}^3$, during the early season, and the maximum errors are in the south-west corner of the field where there are no \textit{in situ} training sites. This is also observed during the late season bare soil conditions in Figure~\ref{fig:354}. For the sweet-corn land cover, shown in Figure~\ref{fig:135}, and the cotton land cover, shown in Figure~\ref{fig:156}, the error in SM is the highest at the boundaries of the field and the south west corner of the field. On DoY 222, shown in Figure~\ref{fig:222}, even when there was maximum heterogeneity in LC with corn, cotton, and bare soil, the errors are minimal, although the standard deviation of the error is high at 0.027 $\mathrm{m}^3/\mathrm{m}^3$. The time-averaged error across the region was 0.027 $\mathrm{m}^3/\mathrm{m}^3$, with a standard deviation of 0.012 $\mathrm{m}^3/\mathrm{m}^3$. The maximum error of 0.0543 $\mathrm{m}^3/\mathrm{m}^3$ was observed during the vegetated period for the few bare-soil pixels which had fractional vegetation cover. Excluding these pixels and the bare-soil pixels at the end of the season when the LAI and LC contradict each other due to crop remnants, the error is 0.012 $\mathrm{m}^3/\mathrm{m}^3$ . The maximum time-averaged error is moderately higher at $\leq$ 0.01 $\mathrm{m}^3/\mathrm{m}^3$ for the cotton pixels and 0.018 $\mathrm{m}^3/\mathrm{m}^3$ for corn pixels.

\subsection{BRT$_\mathrm{st}$ Downscaling in the Presence of Data Gaps in LST}

Typically, LST would be required at the time of downscaling. Because the BRT$_\mathrm{st}$ uses historical data, downscaling can be performed in spite of short data gaps in LST. To investigate the robustness of the algorithm during short data gaps in LST, a regression tree is trained without including LST information in the feature vector. For this, Equation~\ref{eq:temporal} is modified as follows,
\begin{align}
f^{t}(x_{nt})=&\mathrm{SM_{nt,insitu}} \nonumber \\ x_{nt} = &([\mathrm{LST}_{(t-D_1)n}^\mathrm{1 km}, \mathrm{LST}_{(t-D_1+1)n}^\mathrm{1 km}, \ldots , \mathrm{LST1}_{tn}^\mathrm{1 km}]\times U,[\mathrm{LAI}_{(t-D_2)n}^\mathrm{1 km}, \mathrm{LAI}_{(t-D_2+1)n}^\mathrm{1 km}, \ldots , \mathrm{LAI}_{tn}^\mathrm{1 km}], \nonumber \\ &[\mathrm{PPT}_{(t-D_3)n}^\mathrm{10 km}, \mathrm{PPT}_{(t-D_3+1)n}^\mathrm{10 km}, \ldots , \mathrm{PPT}_{tn}^\mathrm{1 km}], \mathrm{SM}_{tn}^\mathrm{10 km}, \mathrm{X}_{tn}^\mathrm{10 km}, \mathrm{Y}_{tn}^\mathrm{10 km}]) \nonumber \\
U =& \begin{pmatrix}
 u_{(t-D_1)n} & 0 & \cdots & 0 \\
  0 & u_{(t-D_1+1)n} & \cdots & 0 \\
  \vdots  & \vdots  & \ddots & \vdots  \\
  0 & 0 & \cdots & u_{tn} \\ 
\end{pmatrix} \nonumber \\
\hat{f}^{t}(x_{nt}) = &\mathrm{SM_{nt,downscaled}}
\label{eq:temporal_LST}
\end{align}
where $U = $ is the unavailability matrix whose elements indicate the days on which LST is withheld from the feature-vector $x_{nt}$. For example, if $u_{tn}$ is set to 0, LST on the day of downscaling is not-available. The distinct scenarios investigated here are when   $u_{tn} = 0$, $u_{tn} = u_{(t-1)n} = 0$, $\cdots$, $u_{tn} = u_{(t-1)n} = \cdots = u_{(t-D_1)n} = 0$.

Figure~\ref{fig:gaplst} shows an increase in the average error in downscaled SM from the BRT$_\mathrm{st}$ algorithm when LST is ``unavailable" for on, or upto 3 consecutive days prior to the day of downscaling, for DoY 222 in the year 2008. It also shows an increase in the number of pixels with error $>$ 0.04 $m^3/m^3$. The error increases marginally by $\leq$0.015 $m^3/m^3$ when LST is unavailable on the day of downscaling and one day prior to the day of downscaling. The error increases substantially to 0.05 $m^3/m^3$ when LST is unavailable for 3 days and continues increasing with a slope of almost $0.015m^3/m^3$ when the unavailability of LST is extended to further days. A similar trend is shown by the number of pixels with error in downscaled SM$>$0.04$m^3/m^3$, with the highest increase occurring at 3 days of unavailability. The BRT$_\mathrm{st}$ algorithm is robust to LST being unavailable on the day of downscaling and 2 day prior with upto 95\% of the pixels having an error in downscaled SM of $<$ 0.04 $m^3/m^3$.
\section{Conclusions}
\label{Sec:Conclusion}

In this study, we implemented and evaluated a downscaling methodology based upon the BRT algorithm that utilizes spatio-temporal correlations in SM and other auxiliary variables to estimate fine-scale SM at 1km. This algorithm preserves heterogeneity in fine-scale SM while drastically reducing the amount of \textit{in situ} data needed for training the algorithm. Multiple regression trees were trained using \emph{in situ} SM and RS products, \textit{viz.} PPT,  LST, LAI, and LC on the day of downscaling and the prior year. The best performing regression trees was automatically chosen using ensemble pruning and their outputs were combined using bootstrap aggregation to generate the downscaled estimates. The time averaged error across the region was found to be 0.01 $\mathrm{m}^3/\mathrm{m}^3$, with a standard deviation of 0.012 $\mathrm{m}^3/\mathrm{m}^3$. The robustness of the BRT$_\mathrm{st}$ to simulated ``data gaps" in LST was investigated and data-gaps of upto 3 days, 95\% of the pixels were found to have an error in downscaled SM of $<$ 0.04 $m^3/m^3$. Although the errors in downscaled SM are higher for this case than with all data available, this algorithm provides a reasonably accurate estimate, which cannot be accomplished with any other downscaling algorithm depending on auxiliary data.

It is envisioned that the BRT algorithm evaluated in this study may be applied using satellite-based higher resolution remote sensing data. For example, the PPT data may be obtained from the Global Precipitation Measurement missions and the LAI, LST and LC products are available from the MODIS sensor aboard Aqua and Terra satellites. Historical data for PPT is available from the NASA Tropical Rainfall Measurement Mission satellite since 1997 and data for LAI, LST and LC are available since 2002. The BRT$_\mathrm{st}$ algorithm can utilize these vast data-sets to provide accurate high resolution SM.

\ifCLASSOPTIONcaptionsoff
  \newpage
\fi

\renewcommand{\baselinestretch}{1.0}
\bibliographystyle{IEEEtran}
\bibliography{combined}

%
\clearpage
\label{Tables}
\listoftables
\clearpage
\begin{table}
\renewcommand{\arraystretch}{1.5} 
\centering
\caption{Planting and harvest dates for sweet corn and cotton during the 2007 growing season}
\label{tab:crops}
\begin{tabular}{c|c|c}
\hline
\multicolumn{1}{c|}{Crop} & \multicolumn{1}{c|}{Planting DoY} & \multicolumn{1}{c}{Harvest DoY} \\
\hline
Sweet Corn & 61 & 139 \\
 & 183 & 261 \\
Cotton & 153 & 332 \\
\hline        
\end{tabular}
\end{table}

\clearpage
\begin{table}
\renewcommand{\arraystretch}{1.5} 
\centering
\caption{Days selected for evaluating BRT estimates. These days capture variability in precipitation/irrigation (PPT) and land cover (LC) }
\label{tab:conditions}
\begin{tabular}{r|l|l}                
\hline
\multicolumn{1}{r|}{DoY} & \multicolumn{1}{c|}{PPT} & \multicolumn{1}{c}{LC} \\
\hline
39 & Dry & Bare \\
135 & Dry, Irrigated & Sweet Corn \\
156 & Wet & Cotton \\
222 & Dry, Irrigated & Sweet Corn and Cotton \\
354 & Wet & Bare \\
\hline
\end{tabular}
\end{table}

\clearpage
\begin{table}
\renewcommand{\arraystretch}{1.5} 
\centering
\caption{RMSE, SD, and KL divergence over the 50$\times$50 km$^2$ region for the disaggregated estimates of SM obtained at 1 km using the BRT, SRRM and BRT-TEMPORAL methods.
\newline A - Baresoil pixels with vegetated sub-pixels at 250 m till DoY 332, B - Baresoil pixels after DoY 332
\newline C - Baresoil pixels without any  vegetated sub-pixels at 250 m till DoY 332}
\label{tab:pritable}
\begin{tabular}{c|c|c|c}
\hline
\multicolumn{1}{c|}{Land Cover} & $\mathrm{RMSE_{BRT-750}}$  & $\mathrm{RMSE_{BRT-30}}$ & $\mathrm{RMSE_{BRT-30,TEMPORAL}}$\\
\hline

Corn & $1.8615\times 10^{-17}$ & 0.0735& 0.0022\\
Cotton &$2.4828\times 10^{-04}$ & 0.0910& 0.0053 \\
Baresoil$^\mathrm{A}$ & $5.6222\times 10^{-5}$ & 0.0845& 0.0543\\
Baresoil$^\mathrm{B}$& $5.628\times 10^{-6}$ & 0.1041& 0.0517 \\
Baresoil$^\mathrm{C}$ & $2.5948\times 10^{-6}$ & 0.0452& 0.0012 \\
\end{tabular}
\end{table}

%
\clearpage
\label{Figures}
\listoffigures

\clearpage

\clearpage

\clearpage

\clearpage

\clearpage

\clearpage

\clearpage

\clearpage

\clearpage

\clearpage

\clearpage
\begin{figure}
\centering\noindent\includegraphics[width=\textwidth]{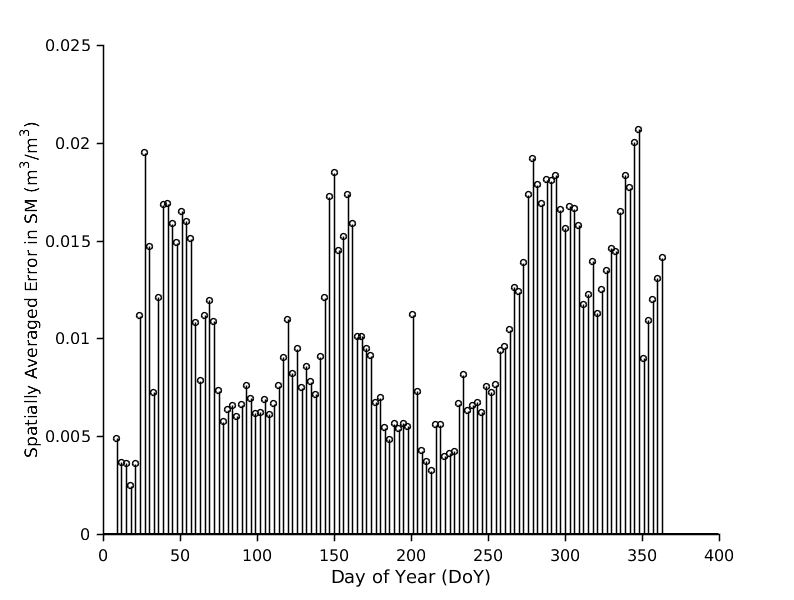}
\caption{Spatially averaged root mean square error in disaggregated soil moisture at 1 km for each day of the year in the validation year when coarse soil moisture is available.}
\label{fig:timeerror}
\end{figure}

\clearpage
\begin{figure}
\centering\noindent\includegraphics[width=\textwidth]{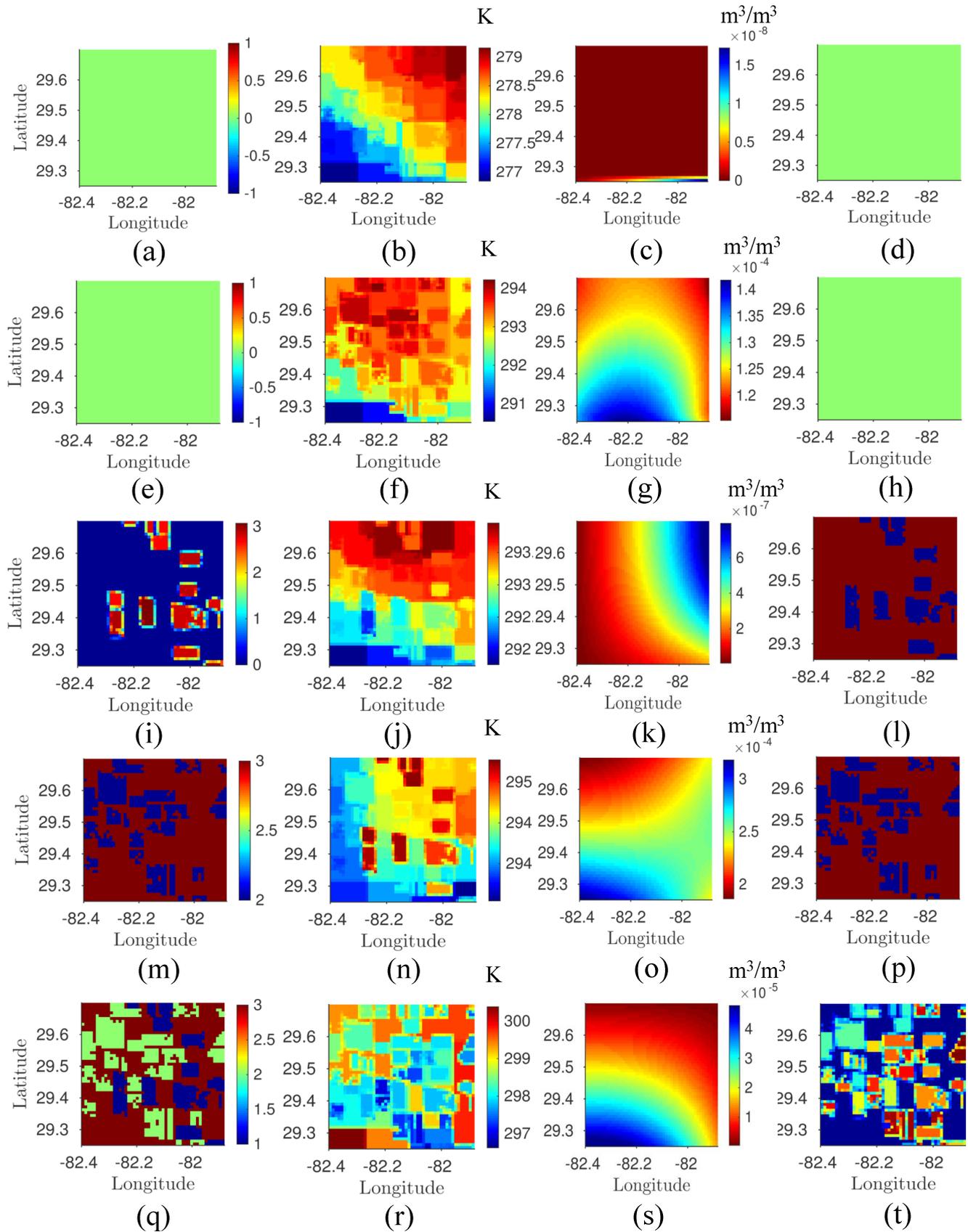}
\caption{Leaf area index, land surface temperature, precipitation and land cover for (a)-(d) DoY 39, (e)-(h) DoY354, (i)-(l) DoY 135, (m)-(p) DoY 156, (q)-(t) DoY 222 respectively. }
\label{fig:daysinputs}
\end{figure}

\clearpage
\begin{figure}
\centering\noindent\includegraphics[width=\textwidth]{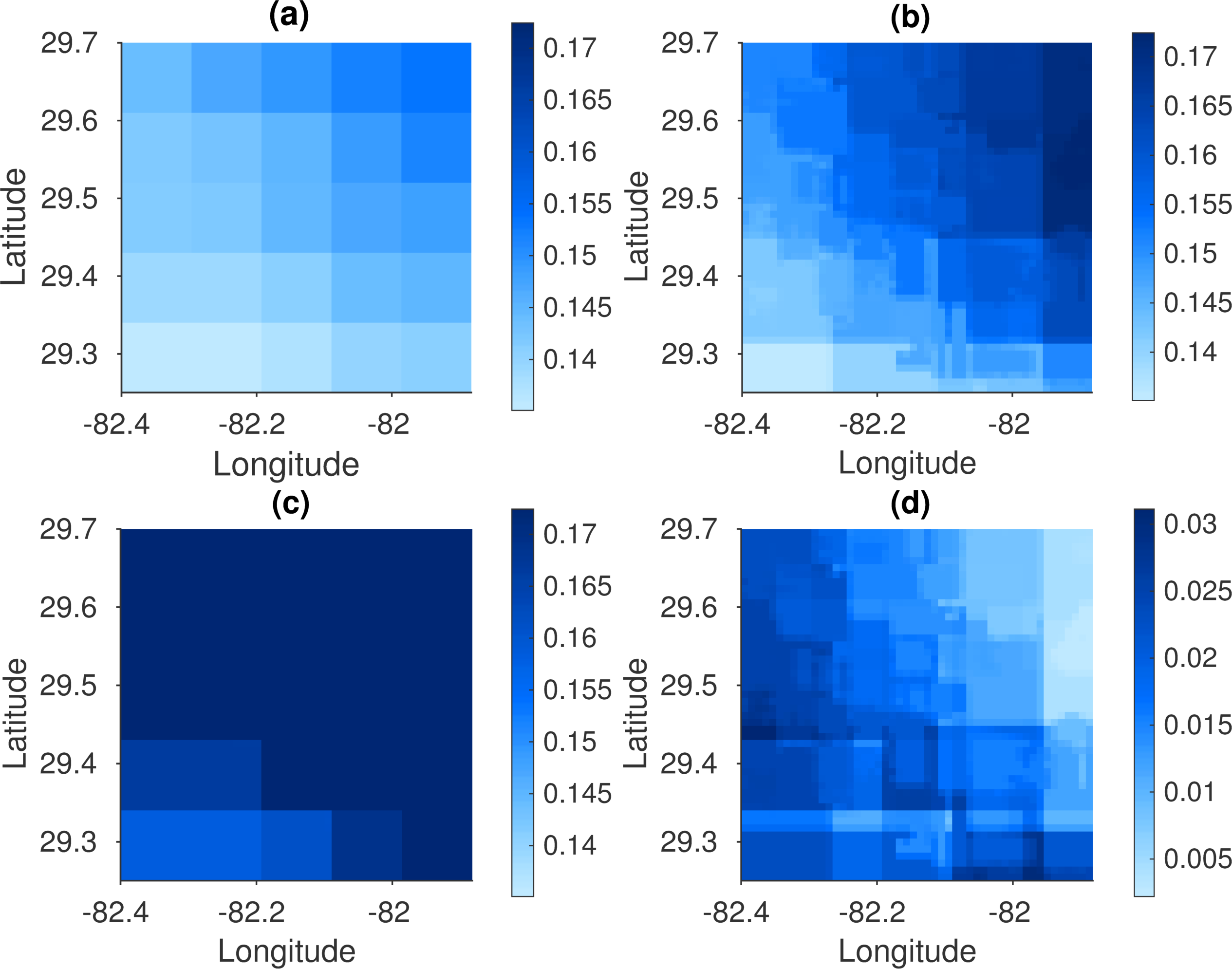}
\caption{DoY 39 - (a) coarse soil moisture at 10 km, (b) true soil moisture at 1 km, (c) downscaled soil moisture using the bagged regression trees algorithm with spatio-temporal training, and, (d) absolute difference between true and downscaled soil moisture.}
\label{fig:39}
\end{figure}

\clearpage
\begin{figure}
\centering\noindent\includegraphics[width=\textwidth]{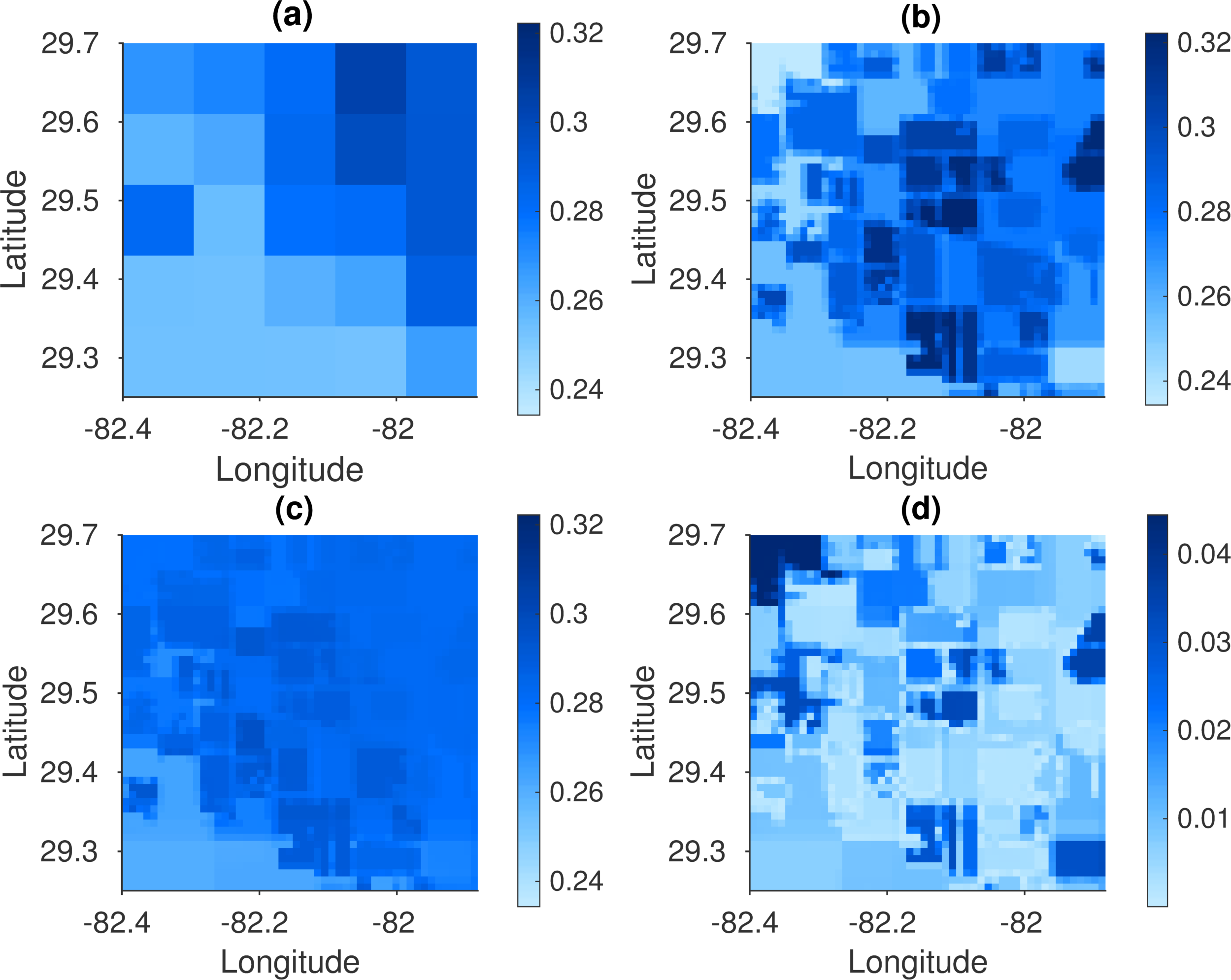}
\caption{DoY 354 - (a) coarse soil moisture at 10 km, (b) true soil moisture at 1 km, (c) downscaled soil moisture using the bagged regression trees algorithm with spatio-temporal training, and, (d) absolute difference between true and downscaled soil moisture.}
\label{fig:354}
\end{figure}

\clearpage
\begin{figure}
\centering\noindent\includegraphics[width=\textwidth]{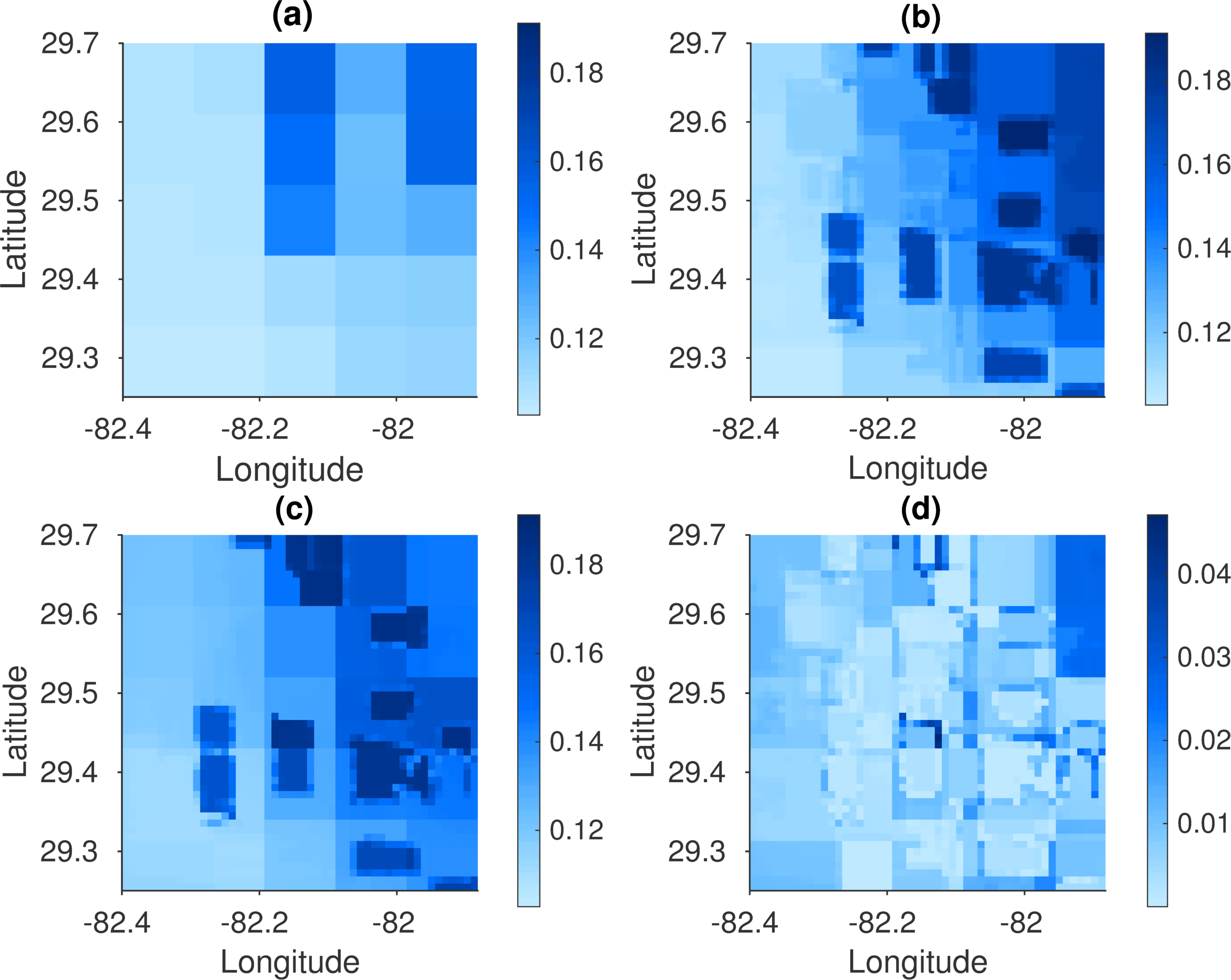}
\caption{DoY 135 - (a) coarse soil moisture at 10 km, (b) true soil moisture at 1 km, (c) downscaled soil moisture using the bagged regression trees algorithm with spatio-temporal training, and, (d) absolute difference between true and downscaled soil moisture.}
\label{fig:135}
\end{figure}

\clearpage
\begin{figure}
\centering\noindent\includegraphics[width=\textwidth]{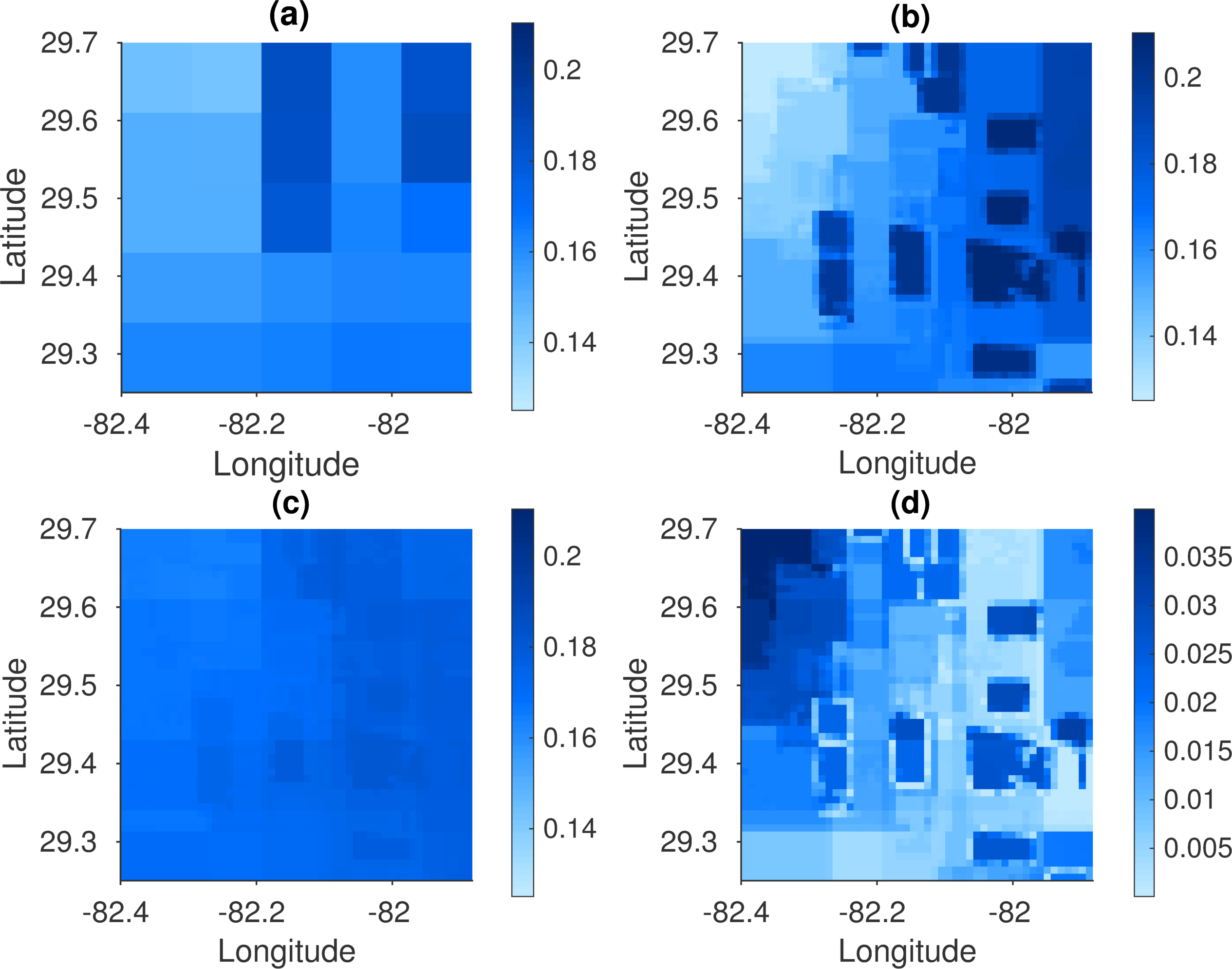}
\caption{DoY 156 - (a) coarse soil moisture at 10 km, (b) true soil moisture at 1 km, (c) downscaled soil moisture using the bagged regression trees algorithm with spatio-temporal training, and, (d) absolute difference between true and downscaled soil moisture.}
\label{fig:156}
\end{figure}

\clearpage
\begin{figure}
\centering\noindent\includegraphics[width=\textwidth]{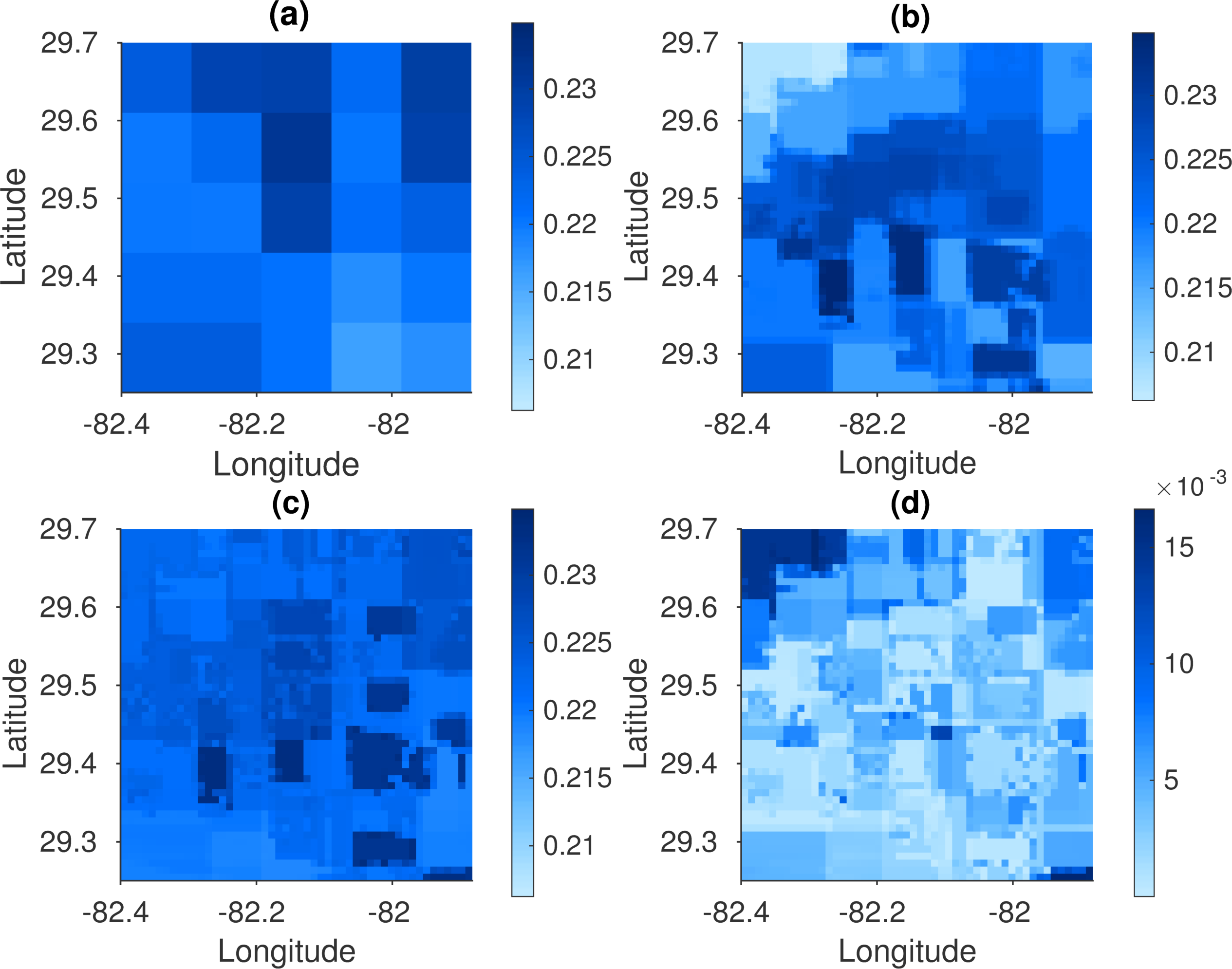}
\caption{DoY 222 - (a) coarse soil moisture at 10 km, (b) true soil moisture at 1 km, (c) downscaled soil moisture using the bagged regression trees algorithm with spatio-temporal training, and, (d) absolute difference between true and downscaled soil moisture.}
\label{fig:222}
\end{figure}

\clearpage
\begin{figure}
\centering\noindent\includegraphics[width=\textwidth]{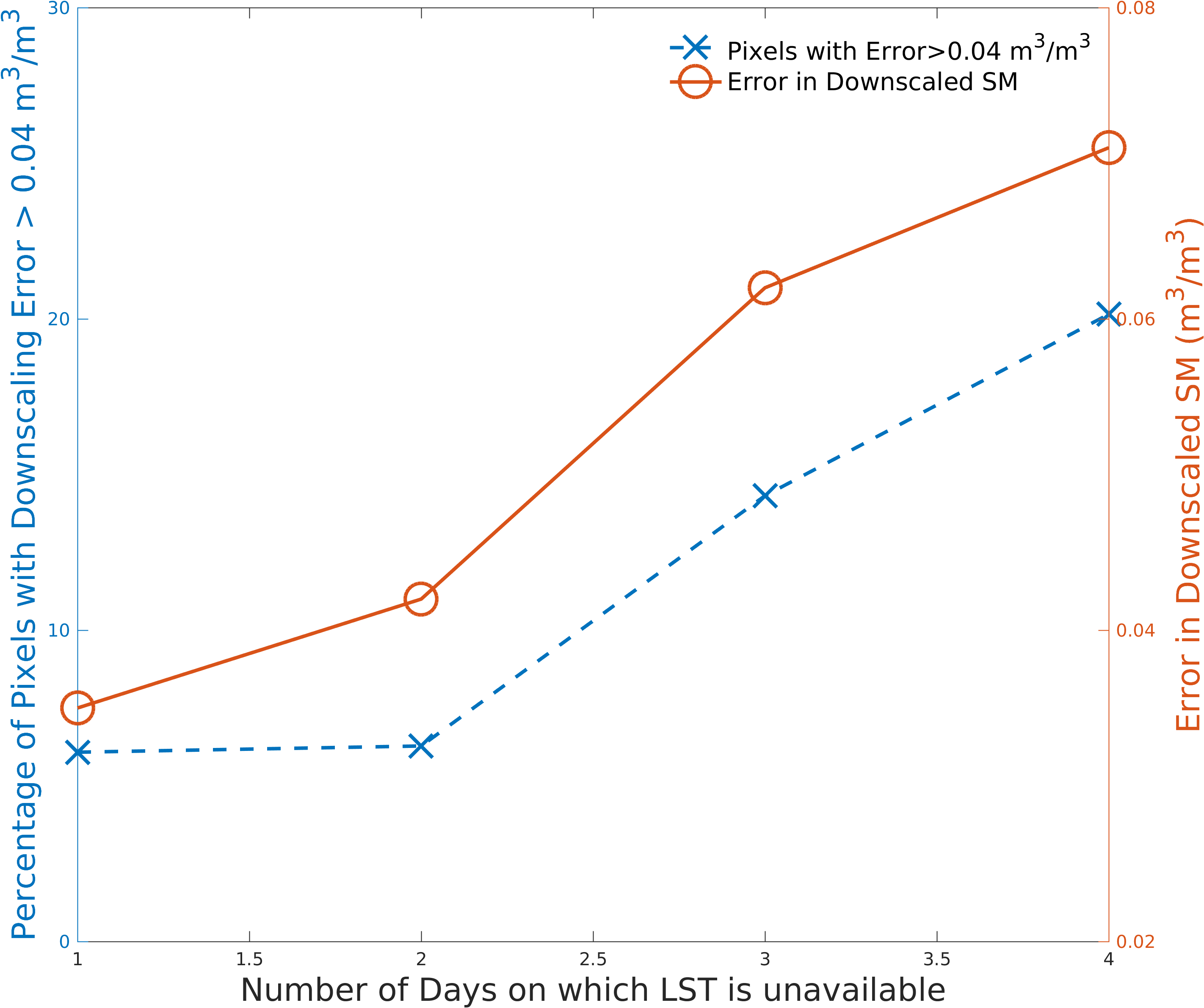}
\caption{Error in downscaled soil moisture as a function of number of gaps in land surface temperature data for day of year 222.}
\label{fig:gaplst}
\end{figure}

\end{document}